\documentclass[letterpaper, 10 pt, conference]{ieeeconf}  

\IEEEoverridecommandlockouts                              

\usepackage{amsmath} 
\usepackage{amssymb}  
\usepackage{pifont}%
\usepackage{flexisym}

\let\labelindent\relax
\usepackage{enumitem}

\usepackage[noadjust]{cite}
\setlist[itemize]{left=0pt,labelindent=0pt,labelsep=0.45em,itemsep=0pt,topsep=2pt}

\usepackage[table,dvipsnames]{xcolor}

\usepackage{graphicx}
\usepackage{diagbox}
\usepackage{booktabs}
\usepackage{multirow}
\usepackage{tabularx}
\usepackage{subcaption}
\usepackage{tikz}
\usepackage{cuted}

\usepackage[table]{xcolor}
\usepackage{graphicx}
\usepackage[normalem]{ulem}

\definecolor{best}{HTML}{FFC0CB} 

\usepackage[breaklinks,colorlinks]{hyperref}

\title{\LARGE \bf
\emph{PhysGraph}: Physically-Grounded Graph-Transformer Policies for Bimanual Dexterous Hand–Tool–Object Manipulation
}

\author{
Runfa Blark Li,
David Kim,
Xinshuang Liu,
Keito Suzuki,
Dwait Bhatt,
Nikola Raicevic,
Xin Lin, \\
Ki Myung Brian Lee,
Nikolay Atanasov,
Truong Nguyen \\
University of California, 
San Diego
}

\begin{document}

\maketitle

\begin{strip}
    \vspace*{-1.5cm}
    \centering
    \includegraphics[width=\linewidth]{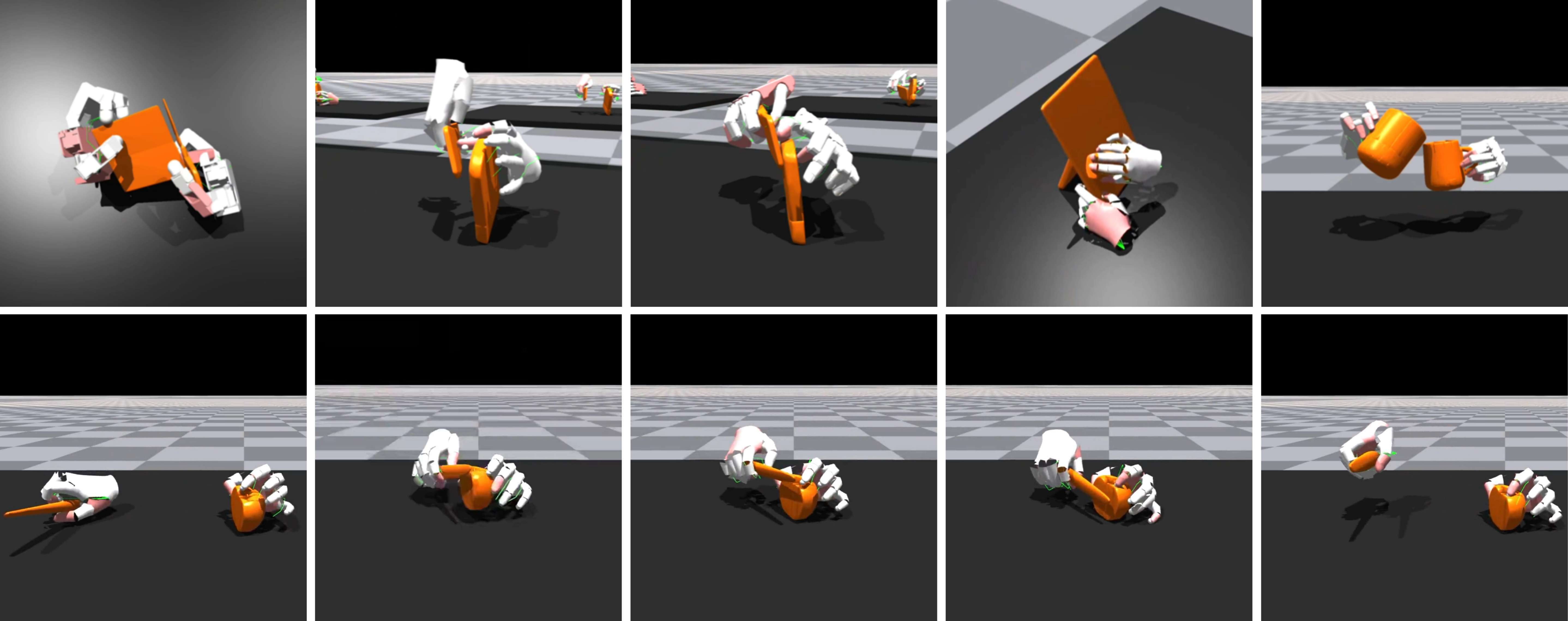}
    \captionof{figure}{\textbf{\emph{PhysGraph}} is a physically-grounded graph transformer policy for bimanual hand-tool-object manipulation. \textbf{Top row}: PhysGraph policy trained and tested on diverse bimanual tool-use tasks with different robot hands (Allegro, ArtiMano, Shadow). \textbf{Bottom row}: Zero-shot policy transfer to unseen tool and object for similar tasks, e.g., trained on ``slicing bread with chop knife'' and tested on ``cutting apple with fruit knife''. Our code and additional results are available at: {\small \url{https://blarklee.github.io/PhysGraph_website_official/}}}
    \label{fig:first_fig}
\end{strip}


\begin{abstract}
Bimanual dexterous manipulation for tool use remains a formidable challenge in robotics due to the high-dimensional state space and complicated contact dynamics. 
Existing methods naively represent the entire system state as a single configuration vector, disregarding the rich structural and topological information inherent to articulated hands. 
We present PhysGraph, a physically-grounded graph transformer policy designed explicitly for challenging bimanual hand-tool-object manipulation. 
Unlike prior works, we represent the bimanual system as a kinematic graph and introduce per-link tokenization to preserve fine-grained local state information. We propose a physically-grounded bias generator that injects structural priors directly into the attention mechanism, including kinematic spatial distance, dynamic contact states, geometric proximity,
and anatomical properties. This allows the policy to explicitly reason about physical interactions rather than learning them implicitly from sparse rewards. 
Extensive experiments show that PhysGraph significantly outperforms baseline - ManipTrans in manipulation precision and task success rates while using only 51\% of the parameters of ManipTrans. Furthermore, the inherent topological flexibility of our architecture shows qualitative zero-shot transfer
to unseen tool/object geometries, and is sufficiently general to be trained on three robotic hands (Shadow, Allegro, Inspire).

\end{abstract}
\section{INTRODUCTION}


Bimanual dexterous manipulation is a long-standing goal in robotics aiming to enable robots to perform human-level multi-contact coordinated manipulation. Although recent learning-based approaches have made substantial progress in dexterous manipulation, bimanual tool-use remains particularly challenging because the robot must maintain stable grasps, regulate contact forces, and coordinate both hands while the tool mediates interaction between the hands and objects. Multi-fingered hands offer superior dexterity and robustness to variations in object geometry, compared to parallel-jaw grippers, but they come with high-dimensional state and action spaces and complex kinematic topology. For this reason, it remains difficult to learn reliable bimanual tool-use policies in practice.

In this paper, we postulate that a key bottleneck is related to how the control policy represents and reasons about the kinematic structure of the hand, tool, and object. Many recent dexterous manipulation policies treat the hand-tool-object states as a flattened single vector \cite{maniptrans, bidexhd, dexmachina, dexman} fed into a multi-layer perceptron (MLP) to implicitly discover (i) the kinematic dependencies within each hand, (ii) cross-hand coordination patterns, and (iii) time-varying interactions among hand, tool, and object. The model must implicitly learn that the ``index fingertip” is kinematically adjacent to the ``index knuckle” from scratch.
However, bimanual tool-use is inherently structured: the system is naturally described by a graph whose nodes (links) and edges (joints/contact) define feasible pathways for information flow and physical influence. Ignoring this structure forces the policy to learn physical reasoning from sparse reward signals, often leading to brittle and physically implausible behaviors in challenging contact-rich tasks.


To address this, we propose PhysGraph, a physically-grounded graph-transformer policy for bimanual hand-tool-object manipulation, illustrated in Figure \ref{fig:first_fig}. Rather than flattening the state, we formulate a graph structure, where nodes represent individual rigid bodies (links, tools, objects) and edges represent physical couplings (joints, contact). Our approach introduces two key innovations: (i) a per-link tokenization strategy and (ii) a physically-grounded head-specific transformer bias generator. First, instead of pooling states into a global embedding, we process each link's states as a distinct token, preserving fine-grained local properties. Second, unlike generic Graph Transformers (e.g., Graphormer \cite{graphormer}) that utilizes abstract graph distances, we inject a learning-based head-specific composite bias directly into the attention mechanism. The composite bias includes \emph{spatial bias} (kinematic chain distance), \emph{edge bias} (joints and contact), \emph{geometric bias} (positions proximity), and \emph{anatomical priors} (serial/synergistic kinematics). These bias terms enable our policy to explicitly reason about the physical connectivity and contact logic, focusing attention on contacting fingers or coordinated joints.
Our main contributions are summarized as follows.

\begin{itemize}
\item We propose PhysGraph, the first graph-transformer policy for high-DoF challenging bimanual dexterous tool-use tasks, which explicitly models the hand-tool-object interactions as a link-joint-contact graph and processes per-link tokenized 
observations.

\item Structural priors are injected into the PhysGraph transformer via a novel physically-grounded bias generator, which provides topological, edge, distance, and anatomical bias via head-specific masking.


\item Extensive experiments on challenging bimanual tool-use tasks demonstrate that PhysGraph significantly outperforms SOTA baseline - ManipTrans \cite{maniptrans}
in success rate and motion fidelity while using only \textbf{\emph{51\%}} of the parameters of ManipTrans, supporting zero-shot transfer to unseen tools/objects, and is compatible with popular robotic dex-hands (Shadow \cite{shadow}, Allegro \cite{allegro}, Inspire \cite{inspire}).

\end{itemize}

\section{RELATED WORK}
\noindent
\textbf{Bimanual Dexterous Manipulation}. Recent works have focused heavily on learning diverse skills from human demonstrations. DexMimicGen \cite{dexmimicgen} and DexMan \cite{dexman} utilize automated data generation and video-based learning to solve basic bimanual tasks like handover and pick-and-place. Similarly, Bi-DexHands \cite{bidexhd} and DexMachina \cite{dexmachina} explore functional retargeting to map human hands key-points to robot hands. \cite{obj-cen} further improves coordination by prioritizing object state representation. ManipTrans \cite{maniptrans} stands out as the SOTA baseline, which employs a residual learning architecture to efficiently transfer human skills to robotic hands. However, these methods typically rely on global states, which often fail to capture the fine-grained dynamics required for the challenging tool-use tasks we address.

\noindent
\textbf{Dexterous Tool-Use \& Sim-to-Real Challenges.} While gripper-based tool use is well-explored, dexterous tool use remains nascent. Concurrent SOTA works have prioritized the Sim-to-Real gap over task complexity. For instance, \cite{tooluse-yip} successfully demonstrates in-hand manipulation of articulated tools on real hardware. However, to achieve the Sim-to-Real, they simplify to hand-tool single-hand task without tool-object interaction. Similarly, \cite{lin} achieves robust Sim-to-Real transfer for humanoid bimanual tasks but only focuses on grasp-and-reach behaviors rather than precise tool-object interaction. DexWild \cite{dexwild} and ClutterDexGrasp \cite{clutterdexgrasp} addressed manipulation in unstructured, cluttered environments using scalable neural control. We take the complementary stance: \emph{Rather than simplifying the task for Sim-to-Real, we focus on solving the high-complexity problem of bimanual tool-object interaction in simulation first} (like precise paper shearing or bread cutting), providing a high-fidelity teacher policy. Our PhysGraph tackles the full complexity of high-DoF bimanual tool-object interaction, establishing a high-performance teacher policy that can serve as a foundation for future Sim-to-Real efforts.

\noindent
\textbf{Learning from References.} A growing trend involves leveraging human video data to synthesize physically plausible motions for humanoids and hands. SkillMimic \cite{skillmimic} recently demonstrated the ability to learn highly dynamic skills - basketball from demonstrations. Similarly, TokenHSI \cite{tokenhsi} introduced a tokenized approach for unifying human-scene interactions, inspiring our own per-link tokenization strategy. While full-body learning demonstrated that humanoid agents can learn diverse interactions often without explicit reference state inputs to the policy, this "reference-free" exploration is insufficient for dexterous hands due to the extremely higher complexity of finger-object contact dynamics. As highlighted by DexMan \cite{dexman}, the high-dimensional action space and dexterity of multi-fingered hands necessitates stronger guidance. Consequently, SOTA dexterous policies, including DexTrack \cite{dextrack}, DexMachina \cite{dexmachina}, OmniGrasp \cite{omnigrasp}, and ManipTrans \cite{maniptrans}, adopt reference-conditioned learning to make training tractable. PhysGraph aligns with this established paradigm by utilizing reference inputs to guide exploration, but distinguishes itself by explicitly modeling the dynamic topology of bimanual tool use through a topology-aware graph architecture rather than a flat vector focusing on single-agent interaction.

\noindent
\textbf{Structured and Graph-Based Policy Learning.} To handle the complexity of high-DoF systems, there is a growing interest toward morphology-aware architectures. The use of Graph Neural Networks for robot control traces back to NerveNet \cite{nervenet}, which explicitly embedded the robot's kinematic graph into the policy network to create morphology-aware controllers.
Graphormer \cite{graphormer} 
later demonstrated the power of attention mechanisms for graph representation. GCNT \cite{gcnt} and GET-Zero \cite{get-zero} recently proposed Graph-Transformer policies to handle variable robot morphologies, proving that graph structures can generalize across embodiments. However, these methods defined their graph edges limited to static and unchangeable locomotion or kinematics, such as the static bones/limbs \cite{gcnt, get-zero} and chemical bonds \cite{graphormer}. PhysGraph builds upon this ``Graph Story" but evolves it for manipulation. We introduce dynamic contact edges and anatomical biases specifically for tool use. While these methods uses static graphs for morphology, PhysGraph dynamically rewires the graph topology based on contact states, allowing the policy to explicitly reason about the tool as a transient extension of the hand’s kinematic chain.
\section{PhysGraph Overview}
\subsection{Problem Statement}

\begin{figure*}[!t]
  \centering
  \includegraphics[width=\textwidth]{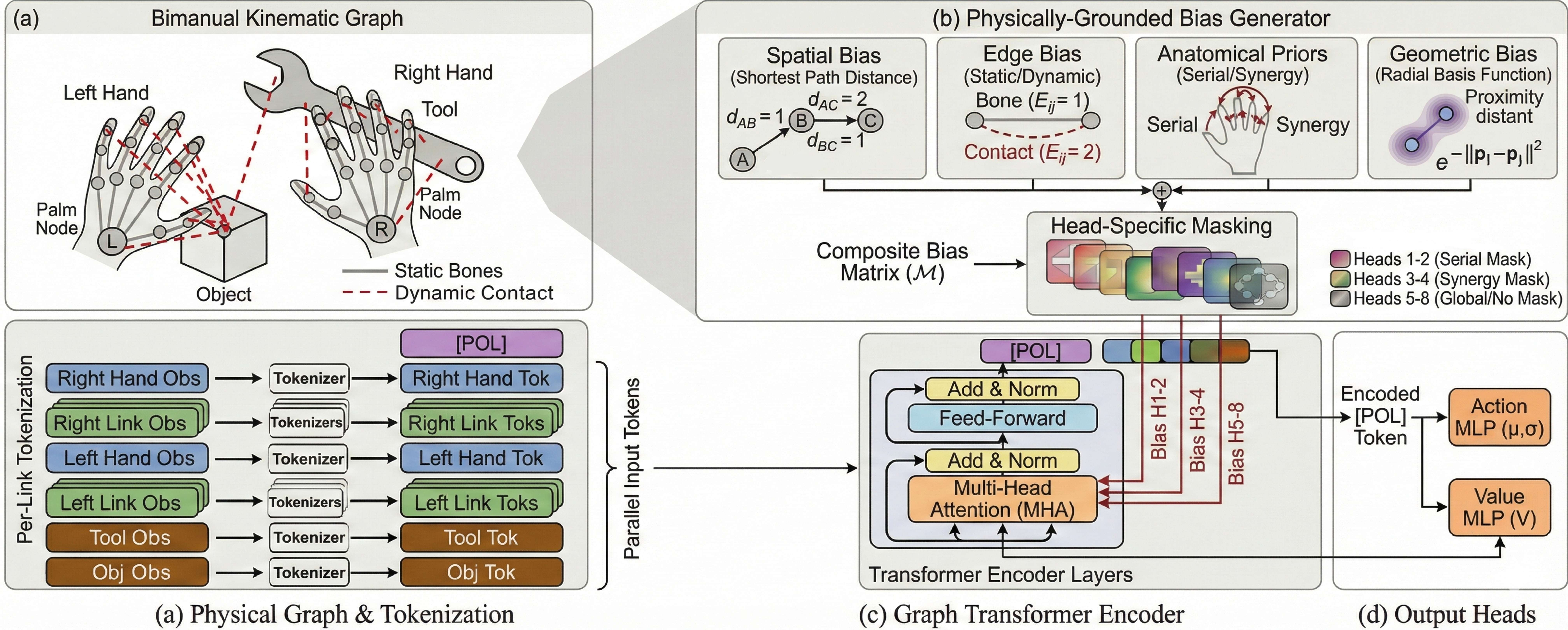}
  \caption{\textbf{Overview of PhysGraph:} 
    \textbf{(a) Physical Graph \& Tokenization:} The bimanual workspace is modeled as a graph where nodes represent links of the left/right hands, tools, and objects and edges represent kinematic (joints) or dynamic (contact) interactions. State-based multi-modal observations for each link are processed into parallel input tokens. 
    \textbf{(b) Physically-Grounded Bias Generator:} This module computes four distinct biases (detailed in Fig. \ref{fig:bias_details}), aggregated into a composite bias matrix $\mathcal{M}$. The biases are applied via head-specific masking, allowing different attention heads to focus on specific physical relationships. 
    \textbf{(c) Graph Transformer Encoder:} The tokenized inputs are processed by a transformer encoder where the Multi-Head Attention (MHA) is modulated by the generated bias in (b). 
    \textbf{(d) Output Heads:} The globally encoded ${[POL]}$ token is passed to MLP heads to predict the policy action distribution ($\mu$, $\sigma$) and value function ($V$).}
    \label{fig:policy_overview}
\end{figure*}

We study bimanual dexterous hand–tool–object manipulation, where two articulated hands must coordinate to interact with a tool and an object under contact-rich dynamics. We model the system as a Markov Decision Process (MDP), defined by a tuple $(\mathcal{S}, \mathcal{A}, p, r, \gamma, T)$, 
where $\mathcal{S}$ is the state space, $\mathcal{A}$ is the action space, $p(s_{t+1} | s_t, a_t)$ is the transition density function induced by rigid-body physics and contact, $r(s,a)$ is the reward function, $\gamma \in (0,1]$ is a discount factor, and $T$ is the horizon.


At each timestep $t$, the agent receives the state $s_t \in \mathcal{S}$
composed of state-based features (proprioception and interaction-related signals), alongside a reference motion state $s^{\text{ref}}_t$ derived from human demonstrations. The use of reference as the goal is important and a standard operation in SOTA works for dexterous policy training \cite{maniptrans, bidexhd, dexman, dexmachina, obj-cen}. The policy outputs actions $a_t \in \mathcal{A}$:

\begin{equation}
\qquad a_t \sim \pi_\theta(\cdot \mid s_t).
\end{equation}

The objective is to learn a policy maximizing the expected discounted return from the initial state distribution, representing our value function: 

\begin{equation}
    V^{\pi_\theta}(\boldsymbol{s}_0) = \mathbb{E}_{\tau \sim \pi_\theta} \left[ \sum_{t=0}^{T-1} \gamma^t r(\boldsymbol{s}_t, \boldsymbol{a}_t) \right],
\end{equation}

%

\noindent
where the expectation is over trajectories $\tau$ of length $T$ sampled from the policy. The reward $r(s_t, a_t)$ incorporates a task-completion term and a tracking term. Since the reference $s^{\text{ref}}_t$ is provided as part of the state, the tracking term computes the deviation between the robot's current pose and the reference, following~\cite{maniptrans}. This formulates the problem as a reference-tracking RL task.
Our contribution is in designing a novel architecture that effectively parameterizes $\pi_\theta$.
\subsection{PhysGraph Architecture}
Figure \ref{fig:policy_overview} summarizes PhysGraph, where we first construct kinematic graph, followed by the graph transformer designed with the physically-grounded composite biases.

\noindent
\textbf{Kinematic Graph Construction.}
We design a policy architecture that models the physical structure of bimanual manipulation as a graph $\mathcal{G} = (\mathcal{V}, \mathcal{E})$, whose nodes $\mathcal{V}$ correspond to rigid bodies, such as hand links, tool and object, and edges $\mathcal{E}$ correspond to kinematic (joints) or dynamic (contact) interactions.



\noindent
\textbf{Tokenization and Transformer Backbone.} 
We propose a novel per-link tokenization approach to preserve the local structure of interaction: contact occurs at hand links, stability depends on distal joints, and error correction propagates along kinematic chains.
Representing rigid-body links as separate tokens enables a transformer to learn structured message passing, e.g., finger links $\leftrightarrow$ tool $\leftrightarrow$ object. Let $\{ x_i \}_{i=1}^N$ denote per-token inputs
extracted from the structured per-link state $s_t$,
where tokens correspond to right-hand state, right-hand per-link states, left-hand state, left-hand per-link states, tool state, and object state. Each per-token input is embedded into a $d$-dimensional space using a tokenizer $f_i$ followed by layer normalization:
\begin{equation}
    h_i^{(0)} = \mathrm{LN}\big(f_i(x_i)\big)\in \mathbb{R}^d,\qquad i=1,\dots,N.
\end{equation}
We prepend a learnable policy token:
\begin{equation}
    h_0^{(0)} = \mathbf{e}_{\text{POL}} \in \mathbb{R}^d,
\end{equation}
and form the token matrix:
\begin{equation}
H^{(0)} = [h_0^{(0)}, h_1^{(0)}, \dots, h_N^{(0)}]^\top \in \mathbb{R}^{(N+1)\times d}.
\end{equation}
%
We employ an $L$-layer transformer encoder that maps $H^{(\ell-1)} \mapsto H^{(\ell)}$ for $\ell = 1, \ldots, L$. For each layer $\ell$ and attention head $h$, the attention logits are computed as:
\begin{equation} \label{eq:att_logits}
A^{(\ell,h)} = \frac{Q^{(\ell,h)}{K^{(\ell,h)}}^\top}{\sqrt{d_h}} + B^{(h)},
\end{equation}
where, $Q^{(\ell,h)}$, $K^{(\ell,h)} \in \mathbb{R}^{(N+1)\times d_h}$ and $B^{(h)} \in \mathbb{R}^{(N+1)\times (N+1)}$ is a \emph{head-specific composite bias}. Attention weights and outputs follow standard transformers \cite{transformer}.


\noindent
\textbf{Physically-Grounded Composite Bias.} We define the composite bias for each head as a weighted sum of physically meaningful components of four biases:
\begin{equation}
\begin{aligned}
B^{(h)}=
\lambda_{\mathrm{sp}} B_{\mathrm{sp}}^{(h)}+
\lambda_{\mathrm{edge}} B_{\mathrm{edge}}^{(h)}+
\lambda_{\mathrm{geo}} B_{\mathrm{geo}}^{(h)} \\+
\lambda_{\mathrm{anat}}^{(h)} B_{\mathrm{anat}}^{(h)}.
\end{aligned}
\end{equation}
All coefficients $\lambda$ are learnable,
so the network can adapt how strongly it relies on each bias and prior. This design provides a useful separation: in \eqref{eq:att_logits}, the content-based dot-product term captures general task-dependent relevance, while $B^{(h)}$ encodes task-agnostic physically-grounded plausibility. Our core contribution is the novel design of the composite bias terms, which is detailed in the next section.

\section{Physically-Grounded Bias}
\begin{figure*}[!t]
  \centering
  \includegraphics[width=\textwidth]{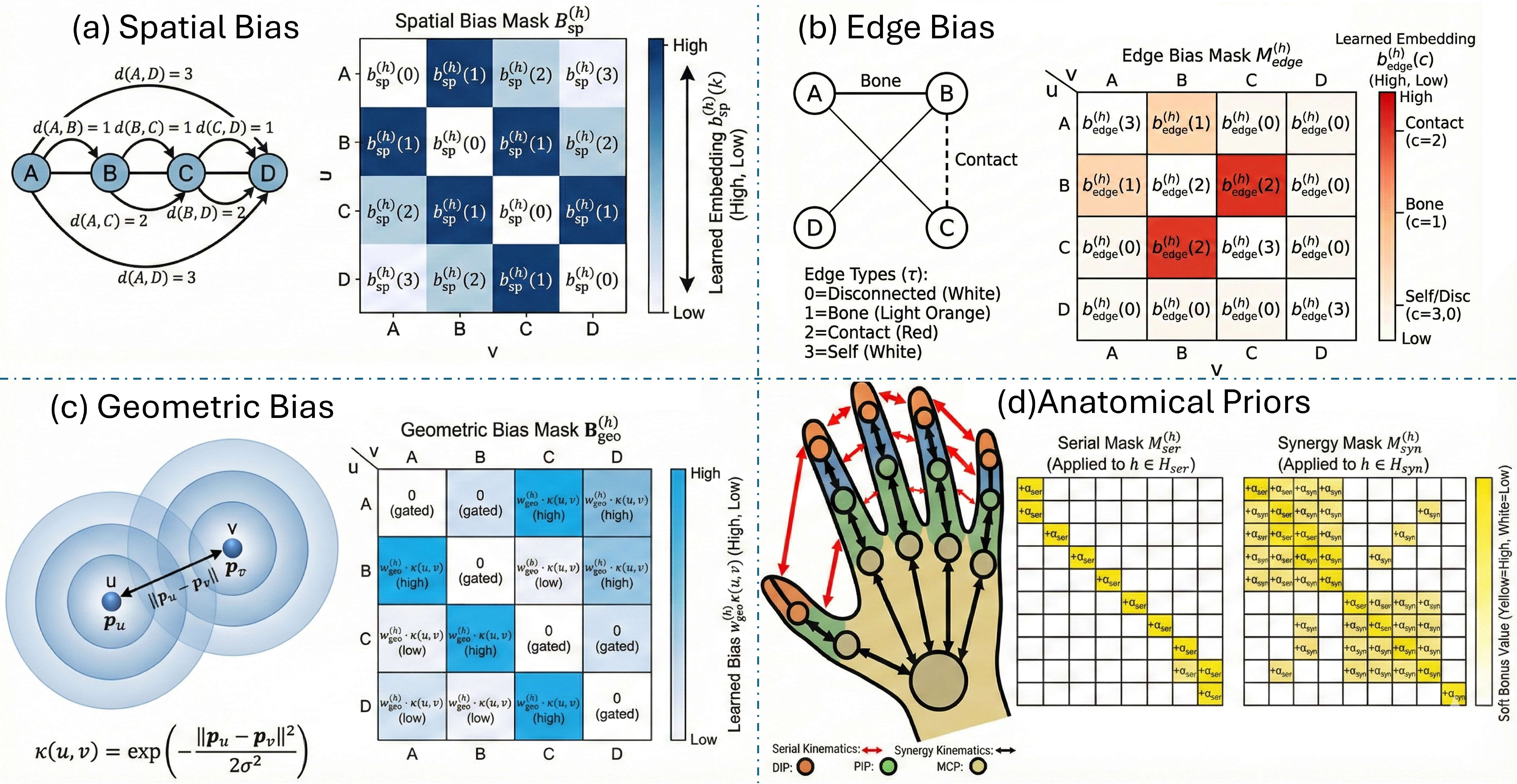}
    \caption{\textbf{Details of the Physically-Grounded Biases.} 
    \textbf{(a) Spatial Bias:} Encodes the topological structure of the hand graph by mapping the shortest path distance $d(u, v)$ between nodes to learned scalar embeddings $b_{sp}^{(h)}$, capturing the ``hop'' distance in the kinematic chain. 
    \textbf{(b) Edge Bias:} Injects structural information by assigning distinct learned embeddings $b_{edge}^{(h)}$ indexed by the edge type $\tau$ connecting two nodes. 
    \textbf{(c) Geometric Bias:} Incorporates spatial proximity in Cartesian space. It utilizes a Radial Basis Function (RBF) kernel $\kappa(u, v)$ to weight attention based on the Euclidean distance $\|\mathbf{p}_u - \mathbf{p}_v\|^2$ between node positions. 
    \textbf{(d) Anatomical Priors:} Encodes knowledge regarding anatomical hand kinematics. The Serial Mask ($M_{ser}$) highlights dependencies along the kinematic chain of a single finger, while the Synergy Mask ($M_{syn}$) promotes attention between corresponding links across different fingers with the same anatomical levels.}
    \label{fig:bias_details}

\end{figure*}

Figure \ref{fig:bias_details} details the four biases: spatial, edge, geometric, and anatomical. Throughout, biases are generated in \emph{node space} on the physical graph $\mathcal{G}$, then lifted to \emph{token space} via a fixed mapping to match the token sequence.

\noindent
\textbf{Spatial Bias (Kinematic Graph Distance).}
Kinematic influence is primarily local along joint chains: distal motion depends strongly on proximal joints, and control corrections often propagate through neighboring links. In sparse-reward RL, asking attention to discover this structure purely from data is inefficient. Spatial bias injects a prior that encourages communication between nodes that are close in the hand kinematic graph.

Let $d(u, v)$ be the shortest graph hop distance between nodes $u$ and $v$ on $\mathcal{G}$. The spatial bias is defined as:
\begin{equation}
    B_{\mathrm{sp}}^{(h)}(u,v) = b_{\mathrm{sp}}^{(h)}(\min(d(u,v), D_{\max})),
\end{equation}
where $b_{\mathrm{sp}}^{(h)}(k)\in\mathbb{R}, k\in\{0,1,\dots,D_{\max}\}$ is a head-specific embedding over discrete hop distances, and $D_{\max}$ is a limit for clipping.
This effectively provides a learnable “graph positional bias” that reflects the kinematic topology. Clipping the distances to $D_{max}$ stabilizes training and prevents the bias vocabulary from growing with graph diameter.

\noindent
\textbf{Edge Bias (Joints + Contact).} Not all node pairs at the same distance are equally important. Some relationships are special: self edges preserve identity, bone adjacency captures rigid kinematic coupling, and contact edges capture instantaneous physical constraints through which forces and torques transmit. Edge bias encodes these semantic edge types directly into attention.

We define discrete edge types $\tau(u,v)$ such that: $\tau(u,v)\in\{0,1,2,3\}$, representing (0) disconnected, (1) bone adjacency, (2) contact
and (3) self. Concretely, using a binary contact indicator $C_t(u, v)$:

\begin{equation}
    \tau(u,v)=
\begin{cases}
3 & u=v, \text{self}\\
2 & C_t(u,v)=1, \text{contact} \\
1 & (u,v)\in\mathcal{E}_{\mathrm{bone}}, \text{adjacency}\\
0 & \text{disconnected}.
\end{cases}
\end{equation}
Using the defined edge type, we define edge bias as:
\begin{equation}
B_{\mathrm{edge}}^{(h)}(u,v) = b_{\mathrm{edge}}^{(h)}(\tau(u,v)),
\end{equation}
where $b_{\mathrm{edge}}^{(h)}(c)\in\mathbb{R}, c\in\{0,1,2,3\}$ is the head-specific edge embedding.
For efficiency, static edge indices can be precomputed once and stored in a buffer, and dynamic contacts can be added as a sparse subset at runtime. This avoids constructing a full $N \times N$ adjacency matrix for the edges, which is important for high throughput. 

During tool-use, contact is the dominant coupling mechanism: Once fingers/palm touch the tool, control depends on integrating local body states, tool/object motion, and response. Our proposed edge bias makes this pathway systematically accessible to attention.

\noindent
\textbf{Geometric Bias (Position Proximity).} Kinematic graph topology alone cannot model interaction locality. Two fingertips from different fingers (or different hands) can become spatially adjacent during grasping even if they are far in the kinematic graph. Similarly, tool–object and hand–tool interactions depend on Euclidean proximity. We propose a geometric bias term that encourages attention to focus on spatially local pairs as a differentiable proxy for physical interaction likelihood.

Let $p_u \in \mathbb{R}^3$ be the position of node $u$. We compute an RBF kernel:
\begin{equation}
\kappa(u,v)=\exp\!\left(-\frac{\|p_u-p_v\|_2^2}{2\sigma^2}\right),
\end{equation}
with learnable bandwidth $\sigma$. We project this proximity scalar into a head-specific bias with learned weight $w_{\mathrm{geo}}^{(h)}$ as:
%
\begin{equation}
\begin{aligned}
    B_{\mathrm{geo}}^{(h)}(u,v) =  \mathbb{I}\big(d(u,v)>d_0\big)\,w_{\mathrm{geo}}^{(h)}\,\kappa(u,v).
\end{aligned}
\end{equation}
The first part of the expression (i.e., $d(u, v) > d_0$) gates geometric bias to pairs that are topologically distant to prevent redundancy with topological bias.

Rigid-body interactions are fundamentally driven by proximity. The RBF term offers a smooth inductive bias: nearby nodes in Cartesian position, regardless of a far kinematics distance, should exchange information, improving responsiveness to emerging contacts and bimanual coordination around the tool and the object.

\noindent
\textbf{Anatomical Priors (Serial + Synergy).} Hands have anatomical structure beyond generic graphs. Each finger forms a kinematic serial chain, while fingers exhibit synergies at the same anatomical level, e.g., coordinated closing/opening and correlated motion of the MCP/PIP joints.
Thus, a single attention head forced to represent all patterns can be inefficient. We instead specialize heads: some focus on serial kinematics, some on synergy couplings, and others remain on global reasoning.

We define two binary relation masks in node space: (i) Serial validity $S(u,v)$: True for near-kinematic-neighbor interactions (e.g., $d(u, v) \le 1$). (ii) Synergy validity $Y(u,v)$: True for anatomically corresponding groups (e.g., fingertip-to-fingertip across fingers; MCP-to-MCP; PIP-to-PIP) within each hand.

We allocate head subsets: $\mathcal{H}_{\mathrm{ser}}$, $\mathcal{H}_{\mathrm{syn}}$,
$\mathcal{H}_{\mathrm{glob}}$.
Instead of hard masking ($-\infty$), we add soft bonuses that encourage—but do not force—these patterns:

\begin{equation}
B_{\mathrm{anat}}^{(h)}(u,v)=
\begin{cases}
\alpha_{\mathrm{ser}}\,S(u,v) & h\in\mathcal{H}_{\mathrm{ser}},\\
\alpha_{\mathrm{syn}}\,Y(u,v) & h\in\mathcal{H}_{\mathrm{syn}},\\
0 & h\in\mathcal{H}_{\mathrm{glob}}.
\end{cases}
\end{equation}

While the hard constraints can harm dexterity, soft bonuses provide an inductive bias while preserving model flexibility. 
\section{EXPERIMENTS}

\begin{table*}[t]
\centering
\caption{\textbf{Quantitative results on bimanual tool-use tasks of Oakink2 dataset \cite{oakinkv2}.} SR: Success Rate; E\_t: Tool \& Object Translation Error; E\_j: Hand Joint Error; E\_ft: Fingertip Error. ArtiMANO Hand. \colorbox{best}{Pink} indicates the best performance.}
\label{table:physgraph_results}
\resizebox{\linewidth}{!}{%
\begin{tabular}{lll l cccc}
\toprule
\textbf{Task ID} & \textbf{Description} & \textbf{Tool (R) / Object (L)} & \textbf{Methods} & \textbf{SR (\%)} $\uparrow$ & \textbf{E\_t (cm)} $\downarrow$ & \textbf{E\_j (cm)} $\downarrow$ & \textbf{E\_ft (cm)} $\downarrow$ \\ \midrule

\multirow{3}{*}{083f7@0} & \multirow{3}{*}{Cut bread with knife} & \multirow{3}{*}{Knife / bread} & Maniptrans & 55.69 & 1.31 & 2.42 & 2.01 \\
 & & & PhysGraph (No Bias) & 82.84 & 1.02 & 2.26 & 1.92 \\
 & & & PhysGraph (full) & \cellcolor{best}90.05 & \cellcolor{best}0.69 & \cellcolor{best}2.17 & \cellcolor{best}1.42 \\ \hline

\multirow{3}{*}{97055@0} & \multirow{3}{*}{Scoop into bowl} & \multirow{3}{*}{Spoon / bowl} & Maniptrans & 74.99 & 1.76 & 3.10 & \cellcolor{best}2.16 \\
 & & & PhysGraph (No Bias) & 82.58 & 1.58 & 2.70 & 2.48 \\
 & & & PhysGraph (full) & \cellcolor{best}86.95 & \cellcolor{best}1.40 & \cellcolor{best}2.38 & 2.24 \\ \hline

\multirow{3}{*}{1292e@0} & \multirow{3}{*}{Pour mug to mug} & \multirow{3}{*}{mug / mug} & Maniptrans & 45.60 & 6.78 & 3.50 & \cellcolor{best}3.35 \\
 & & & PhysGraph (No Bias) & 46.58 & 6.21 & 3.65 & 4.07 \\
 & & & PhysGraph (full) & \cellcolor{best}49.77 & \cellcolor{best}5.93 & \cellcolor{best}3.07 & 4.33 \\ \hline

\multirow{3}{*}{817fb@0} & \multirow{3}{*}{Brush whiteboard} & \multirow{3}{*}{Brush / whiteboard} & Maniptrans & 50.05 & 1.55 & 2.17 & 2.24 \\
 & & & PhysGraph (No Bias) & 58.84 & 1.60 & 1.70 & 1.94 \\
 & & & PhysGraph (full) & \cellcolor{best}62.24 & \cellcolor{best}1.33 & \cellcolor{best}1.54 & \cellcolor{best}1.86 \\ \hline

\multirow{3}{*}{e1fa6@0} & \multirow{3}{*}{Shear paper} & \multirow{3}{*}{Scissors / paper} & Maniptrans & 15.16 & 1.18 & 3.31 & 2.25 \\
 & & & PhysGraph (No Bias) & 27.16 & 1.43 & 2.92 & 1.97 \\
 & & & PhysGraph (full) & \cellcolor{best}35.84 & \cellcolor{best}1.35 & \cellcolor{best}2.52 & \cellcolor{best}1.83 \\ \hline

\multirow{3}{*}{9b728@0} & \multirow{3}{*}{Spread on bread} & \multirow{3}{*}{Spoon / bread} & Maniptrans & 47.22 & 1.05 & \cellcolor{best}3.08 & 1.76 \\
 & & & PhysGraph (No Bias) & 61.69 & 1.01 & 3.20 & 1.86 \\
 & & & PhysGraph (full) & \cellcolor{best}86.64 & \cellcolor{best}0.98 & 3.17 & \cellcolor{best}1.52 \\
\bottomrule
\end{tabular}}
\end{table*}

We conduct extensive experiments to evaluate PhysGraph in learning sophisticated bimanual hand-tool-object manipulation policies from expert demonstrations. In Sec. \ref{sec: arimano}, we benchmark our method against the SOTA baseline on different dynamic, contact-rich bimanual tool-use tasks, and conduct ablation studies to validate the contributions of our proposed physically-grounded biases. In Sec. \ref{sec: zero-shot}, we demonstrate zero-shot transfer to different motion trajectories with unseen tool and object geometries. Finally, we demonstrate PhysGraph’s extensibility across various dexterous hand embodiments in Sec. \ref{sec: multi-emb}.

\noindent
\textbf{Datasets.} For training and evaluation, we utilize the OakInk2 dataset \cite{oakinkv2}. Bimanual manipulation is inherently more complex than single-hand tasks because it requires elaborate coordination between the dominant (tool-using) and non-dominant (object-stabilizing) hands in addition to complicated contact sequences between hand-tool, tool-object, and hand-object. We consider six highly challenging tasks that require sustained dynamic contacts and fine-grained force application. These tasks are described in Table \ref{table:physgraph_results}.

\noindent
\textbf{Metrics.} To evaluate manipulation precision, task compliance, and imitation fidelity, we examine the following metrics: (i) Mean Tool \& Object Translation ($E_t$) and Rotation ($E_r$) Error: The average Euclidean distance (in $cm$) and degree between the 3D positions of the rollout tool/object and the expert reference trajectory, reflecting task compliance; (ii) Mean Hand Joint Error ($E_j$): The mean squared error (in $cm$) across all articulated hand joints, measuring overall kinematic imitation quality; (iii) Mean Fingertip Error ($E_{ft}$): The Euclidean error (in $cm$) specifically evaluating the tracking of the five fingertips, which is critical for assessing contact-rich dexterous manipulation fidelity; (iv) Success Rate (SR): The percentage of trials deemed successful. A trajectory is considered successful only if the above errors remain strictly below specific thresholds. 

\noindent
\textbf{Baselines.} We benchmark PhysGraph against ManipTrans \cite{maniptrans}. 
We leverage ManipTrans as the only baseline because, to the best of our knowledge, ManipTrans achieved the \textbf{best performance} on bimanual dexterous hands manipulation, using an MLP-based residual learning architecture that concatenates all observations into a single state vector.

\noindent
\textbf{Implementation Details.} We use eight transformer heads, where two heads are with the serial mask ($\mathcal{H}_{\mathrm{ser}} \in \{1,2\}$), two heads are with the synergy mask ($\mathcal{H}_{\mathrm{syn}} \in \{3,4\}$), and the rest of the heads are with the global mask ($\mathcal{H}_{\mathrm{glob}} \in \{5,6,7,8\}$). We train the policy using the PPO \cite{ppo} algorithm, and use a PD controller for the forward dynamics. To ensure a strictly fair comparison of the policy architecture, all environmental configurations, reward functions and physical dynamics are kept \emph{identical} to our baseline ManipTrans. 
All experiments run in the Isaac Gym \cite{isaacgym}, utilizing 4096 parallel environments per task at a time step of 1/60 s with an NVIDIA RTX 3090 GPU. For fairness in convergence comparison, all results are extracted from the best-performing checkpoint within a strict maximum of 8k training epochs, which is far more than sufficient to ensure convergence for all methods.

\subsection{Evaluation on Bimanual Tool-Use}
\label{sec: arimano}

\begin{figure}[t]
    \centering
    \includegraphics[width=\columnwidth]{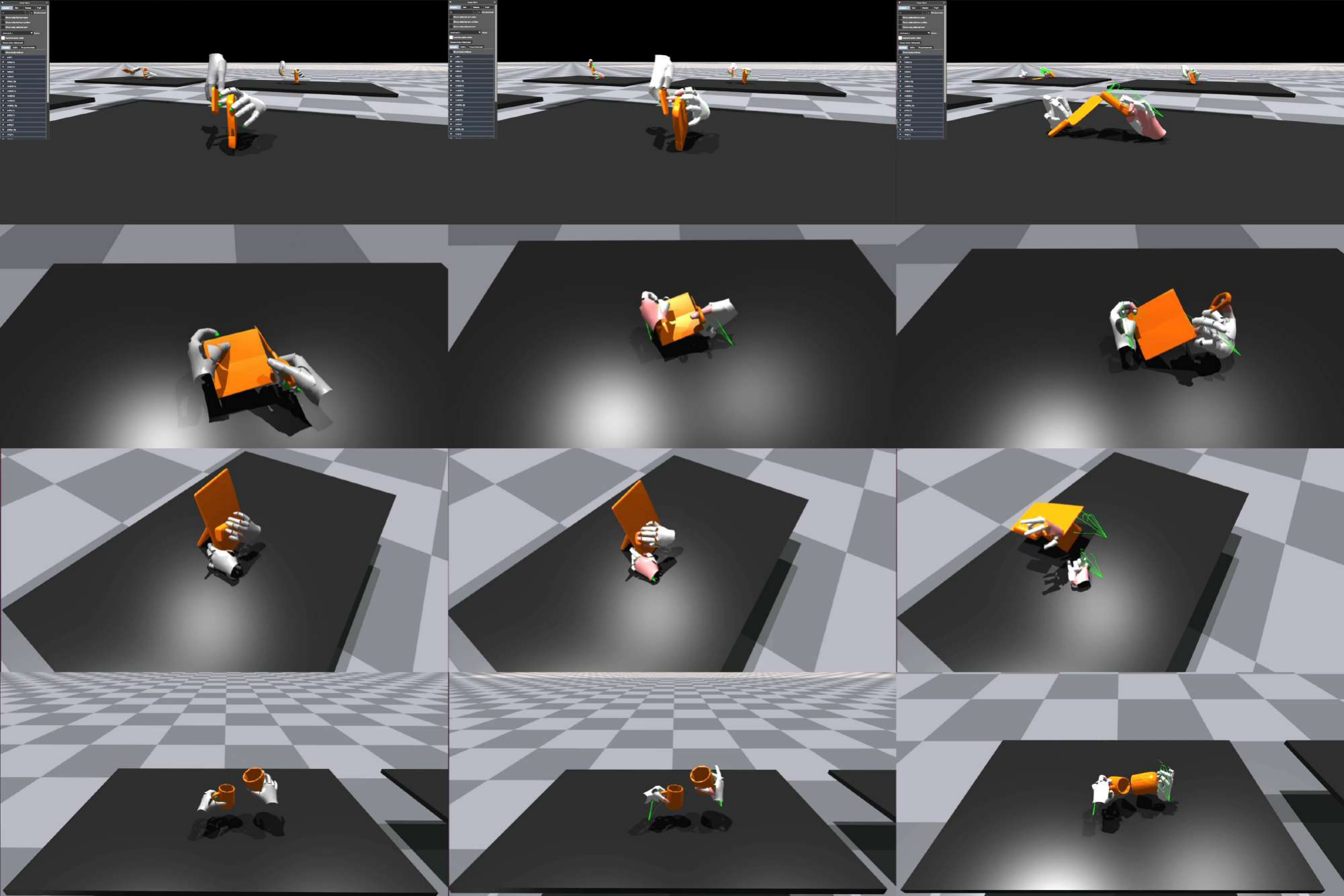}
    \caption{\textbf{Qualitative results on bimanual tool-use tasks of Oakink2 dataset.} Left to right: Ground truth, PhysGraph (ours), ManipTrans. Top to bottom: 0837f@0, e1fa6@0, 817fb@0, 1292e@0. Please refer to the corresponding videos on our website.}
    \label{fig:artimano_all}
\end{figure}

\noindent
\textbf{Quantitative results.} Table \ref{table:physgraph_results} summarizes the evaluation. PhysGraph (Full) achieves consistently higher success rates than ManipTrans across all tasks, with particularly large gains on highly contact-sensitive behaviors such as cutting and spreading. PhysGraph also shows higher fidelity metrics in most settings, indicating more stable interaction trajectories and finer control around contact transitions. Importantly, the number of parameters for the ManipTrans policy for ArtiMANO is \textbf{\emph{3785187}}, whereas the counterpart of PhysGraph is \textbf{\emph{1950792}}. This means PhysGraph exhibits superior performance with much fewer model parameters ($1950792/3785187 = \textbf{51\%}$), which strongly manifests PhysGraph's efficacy and efficiency.

\noindent
\textbf{Qualitative results.} Figure \ref{fig:artimano_all} visualizes the rollouts. Compared to ManipTrans, PhysGraph exhibits more reliable bimanual coordination: ManipTrans frequently exhibits unnatural hand motions, leading to dropped tools (knife, scissors, mug), or misaligned interactions due to implausible force regulation (brush pushes the whiteboard to topple). PhysGraph produces highly natural, closed grasps, with physically plausible force maintaining tight contact throughout the tasks. \textbf{\emph{Please check the corresponding videos at our website and the supplementary material for more details}}.

\noindent
\textbf{Ablation on Physically Grounded Biases}. To validate the contribution of our architectural design, we ablate the physically-grounded bias component  in Table \ref{table:physgraph_results}. Notably, even without explicit physical biases, PhysGraph (No Bias) significantly outperforms the MLP-based ManipTrans across all tasks. This empirically validates the importance of our per-link tokenization strategy: unlike global MLPs that flatten structural information, representing the bimanual system as discrete link tokens preserves local geometric fidelity, enabling superior feature learning even with a standard transformer backbone. However, removing the biases causes a sharp decline in all evaluation metrics. The no-bias variant struggles to maintain dynamic tool-use kinematics, forcing the attention mechanism to discover underlying kinematic structure purely from data, which is highly sample-inefficient. This shows that the proposed physical bias terms provide critical guidance, allowing the heads to communicate along valid physical pathways immediately. The experimental results support our initial motivation: explicitly encoding kinematic topology, edge semantics, geometric proximity, and anatomical insights provides effective inductive biases for contact-rich bimanual tool-use manipulation.

\subsection{Zero-Shot Policy Transfer} 
\label{sec: zero-shot}

\begin{figure}[t]
    \centering
    \includegraphics[width=\columnwidth]{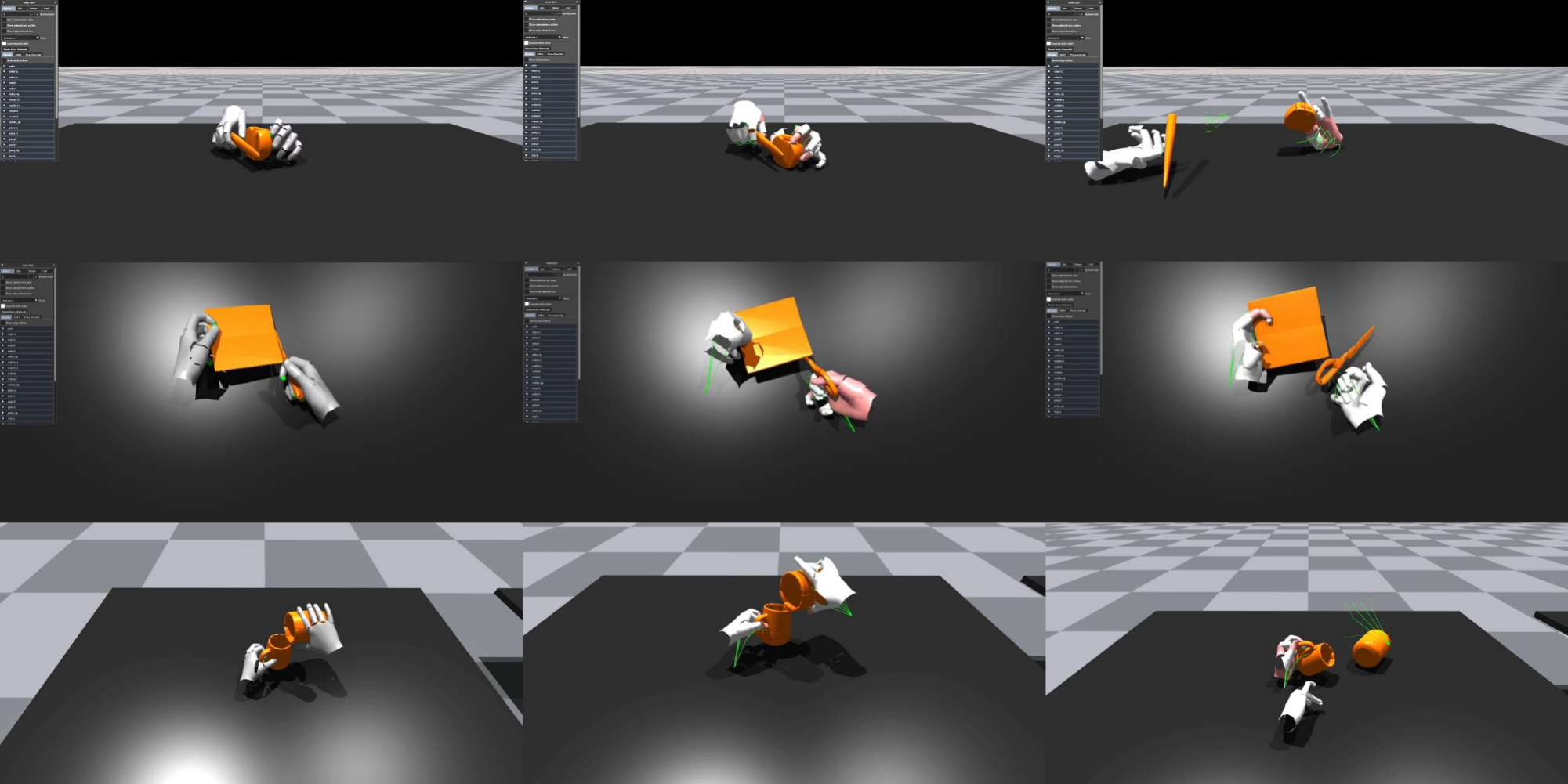}
    \caption{\textbf{Zero-Shot Policy Transfer} \emph{without finetuning}. Left to right: Ground truth, PhysGraph (ours), ManipTrans. Top to bottom (sequence trained $\rightarrow$ sequence deployed): 0837f@0 $\rightarrow$ 9fc3e@0 (chop knife $\rightarrow$ fruit knife; bread $\rightarrow$ apple), e1fa6@0 $\rightarrow$ 66c7f@0, 1292e@0 $\rightarrow$ b9695@0. }
    \label{fig:zero_shot}
\end{figure}

We evaluate PhysGraph's capacity for zero-shot transfer by deploying frozen policies on unseen tool/object for different task instances \textbf{\emph{without finetuning}}. As shown in Fig. \ref{fig:zero_shot}, PhysGraph maintains stronger robustness under instance changes, preserving correct interaction structure with more reliable task completion. As Fig. \ref{fig:zero_shot} shows, a policy trained exclusively to handle a large chop knife and a rectangular bread seamlessly transfers to a much smaller fruit knife and a spherical apple.
This suggests that our physically-grounded bias is inherently spatially invariant; because we explicitly encode relative spatial proximity rather than relying on absolute state coordinates, the model achieves superior generalization beyond the specific instance geometries encountered during training.

\subsection{Embodiment Compatibility Validation}
\label{sec: multi-emb}

\begin{figure}[t]
    \centering
    \includegraphics[width=\columnwidth]{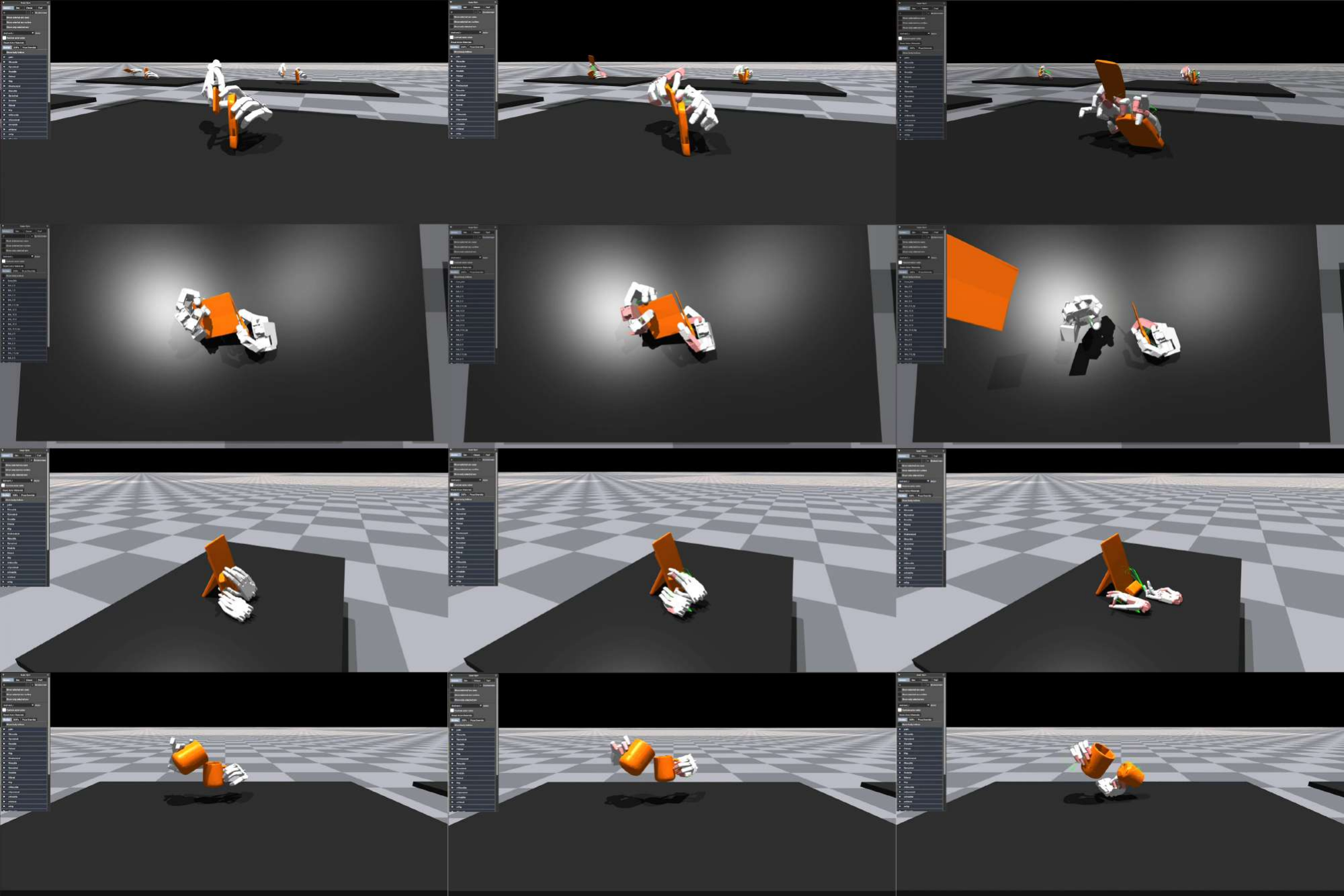}
    \caption{\textbf{Qualitative Results on Multi-Embodiments.} Left to right: Ground truth, PhysGraph (ours), ManipTrans. Top to bottom: 0837f@0 (Shadow), e1fa6@0 (Allegro), 817fb@0 (Shadow), 1292e@0 (Shadow).}
    \label{fig:cross_embodiment}
\end{figure}

\begin{table}[t]
\centering
\caption{\textbf{Quantitative Results of Cross Embodiments.} \colorbox{best}{Pink} indicates the best performance.}
\label{table:embodiment_results}

\setlength{\tabcolsep}{5pt}
\renewcommand{\arraystretch}{1.15}

\begin{tabularx}{\columnwidth}{@{} l l l c c c c @{}}
\toprule
\textbf{Task ID} & \textbf{Hand} & \textbf{Method} & \textbf{SR $\uparrow$} & \textbf{E\_t} $\downarrow$ & \textbf{E\_j} $\downarrow$ & \textbf{E\_ft}  $\downarrow$\\
\midrule

\multirow{6}{*}{083f7@0}
& \multirow{2}{*}{Inspire}  & ManipTrans & 45.07 & 2.02 & 3.83 & 2.92\\
&                              & PhysGraph    & \cellcolor{best}62.14 & \cellcolor{best}1.62 & \cellcolor{best}2.67 & \cellcolor{best}2.3\\
\cmidrule(lr){2-7}
& \multirow{2}{*}{Shadow} & ManipTrans & 22.34 & 1.58 & 3.54 & 2.67 \\
&                              & PhysGraph  & \cellcolor{best}54.77 & \cellcolor{best}1.25 & \cellcolor{best}3.03 & \cellcolor{best}2.26 \\
\cmidrule(lr){2-7}
& \multirow{2}{*}{Allegro} & ManipTrans & 19.67 & 3.60 & 4.56 & 3.17 \\
&                              & PhysGraph   & \cellcolor{best}29.34 & \cellcolor{best}2.95 & \cellcolor{best}4.30 & \cellcolor{best}2.51 \\
\midrule

\multirow{6}{*}{97055@0}
& \multirow{2}{*}{Inspire}  & ManipTrans & 52.95 & 2.16 & 4.20 & 2.91 \\
&                              & PhysGraph  & \cellcolor{best}78.45 & \cellcolor{best}1.17 & \cellcolor{best}3.24 & \cellcolor{best}1.97 \\
\cmidrule(lr){2-7}
& \multirow{2}{*}{Shadow} & ManipTrans & 18.37 & 2.52 & 3.29 & \cellcolor{best}2.71 \\
&                              & PhysGraph  & \cellcolor{best}32.44 & \cellcolor{best}2.26 & \cellcolor{best}2.04 & 2.84 \\
\cmidrule(lr){2-7}
& \multirow{2}{*}{Allegro} & ManipTrans & 23.37 & 3.51 & 4.86 & 3.25 \\
&                              & PhysGraph  & \cellcolor{best}36.28 & \cellcolor{best}2.85 & \cellcolor{best}4.08 & \cellcolor{best}2.41 \\
\bottomrule
\end{tabularx}
\end{table}

We demonstrate that PhysGraph architecture is compatible with various robotic hand topologies. We train and evaluate PhysGraph across multiple dexterous hand embodiments (Shadow \cite{shadow}, Inspire \cite{inspire}, Allegro \cite{allegro}). Table \ref{table:embodiment_results} shows that PhysGraph consistently improves success rate and interaction fidelity across different embodiments compared to the baseline. Figure \ref{fig:cross_embodiment} shows the qualitative results, 
where PhysGraph outperforms the baseline by maintaining physically plausible contact points and interaction force. Since retargeting from MANO introduces additional mismatch for non-MANO embodiments, these results also indicate robustness to realistic trajectory noise and embodiment-specific discrepancies.

\noindent
\textbf{Fingertip Error ($E_{ft}$) Anomaly}. In Tables \ref{table:physgraph_results} and \ref{table:embodiment_results}, we observe that on certain tasks (e.g., pour bottle and shear paper), PhysGraph yields a higher fingertip error $E_{ft}$ than ManipTrans. This is due to a limitation in the ground truth demonstration data rather than policy failure. We used \cite{anyteleop} to extract expert trajectories following ManipTrans~\cite{maniptrans}, aligning human keypoints to the MANO model and retargeting them to the robotic embodiments. However, this retargeting process inherently introduces kinematic noise and a loss of contact fidelity, frequently resulting in non-physical inter-penetrations between the robotic fingertips and the tool/object (e.g., fingertips penetrating whiteboard and mug in Fig. \ref{fig:cross_embodiment} and video). ManipTrans blindly overfits to these erroneous target poses to minimize $E_{ft}$ due to a lack of physical understanding. On the other hand, PhysGraph’s physically-grounded biases penalize non-physical states, causing the policy to deviate from the noisy ground-truth keypoints to discover physically plausible actions. An example is shown in the shear paper task (Fig. \ref{fig:artimano_all}, second row): the retargeted demonstration shows the thumb holding the scissor handles from the outside, which is implausible for human, However, PhysGraph successfully discovers a stable grasp where the thumb correctly inserts through the handle like human. By deviating from the flawed demonstration to respect physical constraints, PhysGraph achieves a higher SR despite the drop in $E_{ft}$. 

\subsection{Limitations and Future Work}
Similar to ManipTrans and other SOTA dexterous manipulation methods \cite{bidexhd,dexmachina,obj-cen,dexman}, the key limitation of PhysGraph is that the policy requires a reference trajectory at inference time. While conditioning the policy on reference stabilizes the learning of high-dimensional control actions with complex contact dynamics, it also constrains deployment when reference trajectories are unavailable. Future work should remove or reduce reliance on reference demonstrations.

\section{CONCLUSIONS}
We proposed PhysGraph for bimanual dexterous hand–tool–object manipulation. Our results suggest that explicit physical structure is crucial for learning challenging contact-rich skills. More broadly, this work highlights that transformer policies for articulated physical systems can benefit from tokenization and attention mechanisms that better reflect their underlying topological, kinematic, and dynamic structure.


\bibliographystyle{IEEEtran}
\bibliography{main}

\end{document}